\documentclass[journal]{IEEEtai}

\usepackage[colorlinks,urlcolor=blue,linkcolor=blue,citecolor=blue]{hyperref}

\usepackage{color,array}

\usepackage{graphicx}
\usepackage{makecell}
\usepackage{xcolor}
\usepackage{orcidlink}
\usepackage{multirow}
\usepackage[T1]{fontenc}
\usepackage{xurl}

\begin{document}
\title{A Survey on Symbolic Knowledge Distillation of Large Language Models}

\author{Kamal Acharya\orcidlink{0000-0002-9712-0265}, \IEEEmembership{Graduate Student Member, IEEE}, Alvaro Velasquez\orcidlink{0000-0001-6757-105X}, \IEEEmembership{Member, IEEE} and Houbing Herbert Song\orcidlink{0000-0003-2631-9223}, \IEEEmembership{Fellow, IEEE}
\thanks{Manuscript received January 6, 2024. This work was supported in part by the U.S. National Science Foundation under Grant No. 2309760 and Grant No. 2317117.}
\thanks{K. Acharya and H. Song are with the Security and
Optimization for Networked Globe Laboratory (SONG Lab), Department of Information Systems, University of Maryland, Baltimore County, Baltimore, MD 21250 USA (e-mail: kamala2@umbc.edu; h.song@ieee.org).}
\thanks{A. Velasquez is with the Department of Computer Science, University of Colorado, Boulder, CO 80309 USA (e-mail: alvaro.velasquez@colorado.edu).}
}

\markboth{IEEE Transactions on Artificial Intelligence, Vol. 00, No. 0, Month 2020}
{Acharya \MakeLowercase{\textit{et al.}}: A Survey on Symbolic Knowledge Distillation of Large Language Model}

\maketitle

\begin{abstract}
This survey paper delves into the emerging and critical area of symbolic knowledge distillation in Large Language Models (LLMs). As LLMs like Generative Pre-trained Transformer-3 (GPT-3) and Bidirectional Encoder Representations from Transformers (BERT) continue to expand in scale and complexity, the challenge of effectively harnessing their extensive knowledge becomes paramount. This survey concentrates on the process of distilling the intricate, often implicit knowledge contained within these models into a more symbolic, explicit form. This transformation is crucial for enhancing the interpretability, efficiency, and applicability of LLMs. We categorize the existing research based on methodologies and applications, focusing on how symbolic knowledge distillation can be used to improve the transparency and functionality of smaller, more efficient Artificial Intelligence (AI) models. The survey discusses the core challenges, including maintaining the depth of knowledge in a comprehensible format, and explores the various approaches and techniques that have been developed in this field. We identify gaps in current research and potential opportunities for future advancements. This survey aims to provide a comprehensive overview of symbolic knowledge distillation in LLMs, spotlighting its significance in the progression towards more accessible and efficient AI systems.
\end{abstract}
\begin{IEEEImpStatement}
There is burgeoning interest in the potential of symbolic knowledge to enhance the interpretability, efficiency, and application scope of LLMs, transforming them into more robust, understandable, and versatile tools. Despite the recognition of its importance, there remains a notable dearth of comprehensive research that thoroughly examines and evaluates the process and implications of this integration. Existing literature predominantly focuses on either the advancements in LLMs or content of the knowledge in the LLMs , with less emphasis on the symbolic knowledge distillation of LLMs. This survey aims to fill this critical gap by offering an extensive review of the current state of symbolic knowledge disitllation in LLMs by highlighting the methodologies, challenges, and advancements in this field.
\end{IEEEImpStatement}


\section{Introduction}
\IEEEPARstart{L}{arge} Language Models (LLMs) are a prominent topic in Artificial Intelligence(AI), with significant breakthroughs occurring frequently. Trained on extensive data sets including websites, research papers, and books, LLMs encapsulate knowledge within their numerous parameters. They can serve as knowledge bases\cite{petroni2019language}, from which information can be extracted and formatted for various purposes, such as fine-tuning other models for specific tasks\cite{west2021symbolic}, validating actions\cite{sclar2022referee}, or generating larger and more accurate datasets\cite{gulcehre2023reinforced}. However, the knowledge embedded in LLMs is not immediately accessible and requires careful extraction and efficient utilization to yield effective results.

The knowledge within LLMs, stored in the weights of their parameters, can be converted into a more interpretable symbolic form through the process of symbolic knowledge distillation. The core challenge here lies in translating the implicit, distributed knowledge encoded in the neural networks of LLMs into explicit, symbolic representations. This transformation is essential for several reasons: to improve the transparency and interpretability of the models, to facilitate knowledge transfer to smaller, more efficient models, and to enable more robust and explainable AI systems. By converting the knowledge into symbolic form, it becomes possible to understand the reasoning behind the model's decisions. This is crucial for applications where understanding the 'why' behind predictions or recommendations is as important as the outcomes themselves. The process is fraught with complexities, including preserving the nuance and depth of the learned knowledge while making it comprehensible and utilizable in a symbolic format.

  In this paper, we introduce a detailed framework dedicated to symbolic knowledge distillation of LLMs, initiating our discussion with a historical overview of symbolic knowledge distillation and its evolutionary path to its current state. Following this, we delve into an analysis of various traditional knowledge distillation methods and their comparison with symbolic knowledge distillation approaches. We further explore LLM architectures, including their training and fine-tuning mechanisms. We classify symbolic knowledge distillation techniques into three distinct categories: Direct, Multilevel, and Distillation via Reinforcement Learning. Additionally, we have compiled research papers focused on symbolic knowledge, as well as those specifically addressing symbolic knowledge distillation of LLMs. Our survey provides a thorough examination of the latest developments in symbolic knowledge distillation of LLMs, highlighting the methodologies, challenges, and progress in the field, thereby offering valuable insights for the research community interested in further exploration of this domain.

The rapid expansion of LLMs has led to the production of numerous survey papers. All the previous survey papers on LLMs cover different aspects except for the symbolic knowledge. Further exploring we find that no survey paper has been published related to the symbolic knowledge distillation. The focus areas of existing survey papers on LLMs include:
\begin{itemize}
    \item Comprehensive overviews of LLMs\cite{zhao2023survey,min2023recent,hadi2023large}
    \item Evaluation of LLMs\cite{chang2023survey}
    \item Code generation\cite{zan2023large}
    \item LLMs in education\cite{KASNECI2023102274}
    \item LLM as Knowledge Base\cite{alkhamissi-lms-as-kbs-22,razniewski2021language}
    \item Reasoning Knowledge in LLMs\cite{huang2022towards}
    \item Explainability in LLMs\cite{zhao2023explainability}
    \item Aligning LLMs with human\cite{wang2023aligning}
    \item Instruction tuning for LLM\cite{zhang2023instruction}
    \item Model Compression in LLM\cite{zhu2023survey}
    \item Trustworthiness evaluation of LLM\cite{liu2023trustworthy}
    \item LLM for software engineering\cite{fan2023large}
    \item Hallucination in LLM\cite{zhang2023siren}
    \item Multimodal LLM\cite{wu2023multimodal}
    \item LLMs for Robotics\cite{zeng2023large}
    \item LLMs for Information Retrieval\cite{zhu2023large}
\end{itemize}
Our work stands in contrast to existing approaches in several key aspects. While traditional methods primarily focus on either the performance enhancement of smaller models or the interpretability aspect of knowledge distillation, our framework synergizes these objectives.

The remainder of this paper is structured as follows: Section II reviews the milestones in knowledge distillation and LLM, establishing the context and background for our work. Section III details the preliminaries about symbolic knowledge distillation and LLM, followed by Section IV, which presents a thorough process of symbolic knowledge distillation in LLM. Section V discusses the related research work that has been carried out. In Section VI, we discuss opportunities that have emerged from Symbolic Knowledge Distillation. Section VII is devoted to the challenges of implementing proposed Symbolic knowledge distillation applications. We identify the obstacles and challenges that may arise. Section VIII highlights the Lesson Learned and Key Takeaways and finally, in Section IX, we offer concluding remarks on our survey paper.


\section{Milestones in Knowledge Distillation and Large Language Models}

Over the last seven decades, language technology has advanced significantly. The Turing Test\cite{turing2009computing}, conducted in 1950, was one of the earliest milestones in this field, which laid the foundation for the concept that machines can perform at the level of humans and demonstrate the intelligence. In the same year Shannon used concept of entropy and provided the way of prediction of the next letter when the preceding text is known\cite{shannon1951prediction}.  In 1964, ELIZA\cite{weizenbaum1966eliza} was introduced as a Natural Language Processing (NLP) computer program which was designed to mimic the conversational style of a psychotherapist. SHRDLU\cite{winograd1971procedures}, introduced in 1968, was an early example of an interactive natural language understanding system which can understand and respond to natural language commands related to a simplified world of objects. Following year was the dominance of the Statistical Language Model(SLM). Notable works that lead the way were "Introduction of Stochastic Approach for Parsing"\cite{sampson1986stochastic} in 1986 and "Statistical Approach to machine translation"\cite{brown1990statistical} in 1990. Due to the problem like  Brittleness Across Domains,  False Independence Assumption and Shannon-Style Experiments, there was downfall of the SLMs\cite{rosenfeld2000two}.

With the introduction of Long Short-Term Memory(LSTM)\cite{hochreiter1997long} in 1997, we entered into the era of Neural Language Model(NLM). These models helped in language processing by capturing the long term dependencies and successfully handling the vanishing gradients. In 2001, the first neural language model was introduced which can be trained using Stochastic Gradient Descent(SGD) algorithm and proved to be computationally efficient and scalable to larger dataset.\cite{bengio2000neural}. Neural Networks not only increased in scope and functionality but also in terms of the size\cite{idelbayev2021lc}. The concept of model compression\cite{buciluǎ2006model} was introduced in 2006. Model compression and acceleration techniques was divided into four different approaches\cite{cheng2018model}: parameter pruning and sharing\cite{wu2016quantized}\cite{courbariaux2015binaryconnect}\cite{sindhwani2015structured}\cite{han2015learning}\cite{wang2018packing}, low-rank factorization\cite{yu2017compressing}\cite{denton2014exploiting}, transferred/compact convolutional layers\cite{cheng2017survey} and knowledge distillation\cite{hinton2015distilling}.

In 2011, IBM Watson made significant strides in language processing by winning a Jeopardy game against human competitors\cite{high2012era}. Two years later, in 2013, the Word2Vec algorithm\cite{mikolov2013efficient} was introduced, which enabled computers to understand the context of a word and its relationship with other words using dense vector representation where similar words are located close to each other. In 2014, seq2seq\cite{sutskever2014sequence} was introduced which used encoder to represent variable length input sequence into fixed length vector and decoder to generate output sequence. In the same year, Global Vectors for Word Representation(GloVe)\cite{pennington-etal-2014-glove} was introduced, which used co-occurance matrix to capture relationship between the words in corpus and was successful in capturing the local and global context informaiton. Knowledge distillation is a model compression technique introduced in 2015 that transfers knowledge from a high-capacity teacher model to a more compact student model. Later in that year FitNets\cite{romero2014fitnets} was introduced that add an additional term along with the knowledge distillation loss. In 2016, study\cite{zagoruyko2016paying} instead of utilizing representations from a specific point in the network, employed attention maps as hints, comparing the mean squared error (MSE) between the attention maps of the student and teacher models. In same year, SQuAD (Standford Question Answering Dataset)\cite{rajpurkar2016squad} was introduced, which facilitated the development of question-answering systems by being benchmark dataset for evaluating machine reading comprehension.

In 2017, the Transformer\cite{vaswani2017attention} model was introduced, which enabled the development of advanced language models that can learn relationships between words in a sentence more efficiently by using the concept of self-attention. In the following year, 2017 \cite{yim2017gift} employed a similar approach. However, instead of utilizing representations or attention maps, they provided hints by using Gram matrices. In 2018, a supplementary module called the paraphraser\cite{kim2018paraphrasing} is incorporated into the model.  In same year, ELMo (Embedding from Language Model)\cite{peters-etal-2018-deep}, context dependent representation of word was introduced which uses different embeddings for same word in different context. Universal Sentence Encoder\cite{cer2018universal} was also introduced in same year, which further enhanced language processing by introducing embeddings for sentence representations and can handle multiple languages.

\begin{figure}
\centerline{\includegraphics[width=18.5pc]{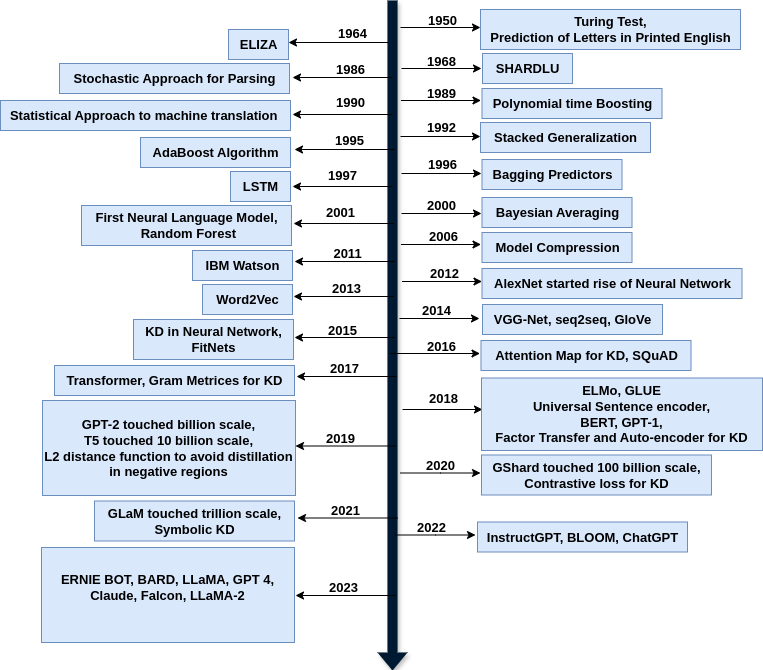}}
\caption{Milestones in history of LLM and Knowledge Distillation}
\label{fig_1}
\end{figure}

General Language Understanding Evaluation(GLUE)\cite{wang2018glue}, a benchmark to evaluate the performance of NLP models on a range of language understanding tasks, became a standard evaluation framework for comparing different language models. Bidirectional Encoder Representations from Transformers(BERT)\cite{devlin2018bert} and Generative Pre-Training-1(GPT-1)\cite{radford2018improving} were introduced in the same year, 2018 which begin the era of Pre-trained Language Model(PLM). In 2019, GPT-2\cite{solaiman2019release} became the first language model to touch a billion scale of parameters. Later that year, T5\cite{raffel2020exploring} became the first language model to touch the 10 billion parameter scale. According to \cite{heo2019comprehensive} published in 2019, the current approach of extracting hints may not be optimal due to the loss of information caused by the ReLU transformation. To address this, they introduced a modified activation function called marginReLU. In \cite{tian2019contrastive} published in 2020, the student model learns from the intermediate representations of the teacher model by employing a contrastive loss over these representations.  As like the way human way of learning, knowledge distillation was applied in the model; self-learning\cite{yuan2019revisit}, mutual learning\cite{zhang2018deep}, teacher student learning\cite{hinton2015distilling}, teacher assistant\cite{mirzadeh2020improved} and continual learning\cite{zhai2019lifelong}. Moreover, the application of knowledge distillation extends beyond transferring knowledge between models. It can also be utilized in various other tasks, including adversarial attacks \cite{papernot2016distillation}, data augmentation \cite{lee2020self}\cite{gordon2019explaining}, data privacy and security \cite{wang2019private}, as well as dataset distillation \cite{wang2018dataset}\cite{bohdal2020flexible}. Between 2010 and 2020, the domain of transfer learning experienced significant expansion, with numerous transfer learning models achieving state-of-the-art results across various disciplines\cite{niu2020decade}.

Google Shard (GShard)\cite{lepikhin2020gshard}, introduced in 2020, became the first language model to touch the 100 billion parameter scale. And in 2021, the Generalist Language Model (GLaM)\cite{du2022glam} became the first language model to touch the trillion parameter scale. Concept of symbolic knowledge distillation\cite{west2021symbolic} was introduced in the same year which is a technique for training smaller models using larger models as teachers and involves distilling knowledge symbolically. Since then symbolic knowledge distillation has been used in various areas such as reference free sentence summarization\cite{sclar2022referee}, comparative knowledge acquisition\cite{howard2023neurocomparatives}.  The scaling laws for neural language models\cite{kaplan2020scaling}, reveal that model performance improves predictably with increases in model size, dataset size, and computational resources, following a power-law relationship. This means that larger models are significantly more efficient in learning from data. In 2022 and 2023, this trend persisted, with various industry leaders introducing new large-scale language models that leveraged these principles to achieve enhanced performance, demonstrating the continued advancement and efficacy of scaling up model size and computational power in the development of language models. Major technology companies are investing heavily in developing their own LLMs because they recognize the immense potential of these systems to revolutionize various industries, such as healthcare, finance, and customer service. Also, LLMs can help these companies maintain their position as leaders in the field of AI and keep up with competitors. Given the swift advancements in this field, there is a pressing need to steer AI towards paths that prioritize safety and responsibility\footnote{\url{https://www.whitehouse.gov/briefing-room/presidential-actions/2023/10/30/executive-order-on-the-safe-secure-and-trustworthy-development-and-use-of-artificial-intelligence/}(last accessed on: [28/02/2024])}. 

The study\cite{hoffmann2022training} concludes that for compute-optimal training, both the model size and the number of training tokens should be scaled equally; specifically, each doubling of the model size should be accompanied by a doubling of the number of training tokens. Conversely, study\cite{villalobos2022will} suggest that the supply of high-quality language data will likely be depleted by 2026. In contrast, low-quality language data and image data are projected to be exhausted between 2030 and 2050 for low-quality language data, and between 2030 and 2060 for image data. The current trajectory of rapidly increasing the parameters of LLMs, which depend on vast datasets, may decelerate unless there are significant improvements in data efficiency or new data sources are discovered. These findings have influenced the development of next-generation LLMs towards models capable of generating their own training data for self-improvement. Furthermore, LLMs will need to incorporate self-fact-checking capabilities. These scenarios underscore the importance of symbolic knowledge distillation and suggest a potential shift of LLMs towards this approach.

It has been utilized for labeling\cite{wang2021want}\cite{smith2022language}, where the teacher model generates outputs based on the provided input, and for expansion\cite{chaudhary2023code}\cite{luo2023wizardmath}, where the teacher model produces samples akin to given demonstrations through in-context learning. For data generation\cite{ding2023enhancing} which involves synthesizing data according to specific meta-information, such as a topic or entity, feedback\cite{jiang2023lion} which involves providing guidance on the student's outputs, encompassing preferences, corrections, and expansions of challenging samples. Finally, for self-checking\cite{yuan2024self}  which entails the student model generating outputs, which are subsequently filtered for high quality or self-evaluated by the student model.

\begin{table*}
\caption{Technical Companies with their LLM}
\centering
\begin{tabular}{|l|l|l|l|l|}
\hline
Companies                     & LLM                                             & Year & Parameters(in billions) & Corpus Size                     \\ \hline
\multirow{11}{*}{Google}      & T5\cite{raffel2020exploring}                                              & 2019            & 11                                & 1 trillion tokens               \\ \cline{2-5} 
                              & GShard\cite{lepikhin2020gshard}                                          & 2020            & 600                               & 1 trillion tokens               \\ \cline{2-5} 
                              & mT5\cite{xue2020mt5}                                           & 2021            & 13                                & 1 trillion tokens               \\ \cline{2-5} 
                              & GLaM\cite{du2022glam}                & 2021            & 1200                              & 1.6 trillion tokens             \\ \cline{2-5} 
                              &   FLAN\cite{wei2021finetuned}                                           & 2021            & 137                               & Not Available                   \\ \cline{2-5} 
                              &   LaMDA\cite{thoppilan2022lamda} & 2022            & 137                               & 1.56T words, 168 billion tokens \\ \cline{2-5} 
                              & Minerva\cite{lewkowycz2022solving}                                         & 2022            & 540                               & 38.5B tokens                    \\ \cline{2-5} 
                              & UL2 \cite{tay2022ul2}                                            & 2022            & 20                                & 1 trillion tokens               \\ \cline{2-5} 
                              & PaLM\cite{chowdhery2023palm}                & 2022            & 540                               & 768 billion tokens              \\ \cline{2-5} 
                              & FLAN-T5\cite{chung2022scaling}                                         & 2022            & 11                                & Not Available                   \\ \cline{2-5} 
                              & FLAN-PaLM\cite{chung2022scaling}                                       & 2022            & 540                               & Not Available                   \\ \cline{2-5}
                              & Gemini(\url{https://gemini.google.com/app}) &2024 &Not Available &Not Available \\
                              \hline
\multirow{7}{*}{OpenAI}       & GPT-2\cite{radford2019language}                                           & 2019            & 1.5                               & 40GB ($\sim$10 billion tokens)  \\ \cline{2-5} 
                              & GPT-3\cite{brown2020language}                                           & 2020            & 175                               & 499 billion tokens              \\ \cline{2-5} 
                              & Codex\cite{chen2021evaluating}                                           & 2021            & 12                                & 100 billion tokens              \\ \cline{2-5} 
                              & WebGPT\cite{nakano2021webgpt}                                          & 2021            & 175                               & Not Available                   \\ \cline{2-5} 
                              & InstructGPT\cite{ouyang2022training}                                     & 2022            & 175                               & Not Available                   \\ \cline{2-5} 
                              & ChatGPT(\url{ https://openai.com/blog/chatgpt })                                    & 2022            & Not Available                     & Not Available                   \\ \cline{2-5} 
                              & GPT-4\cite{openai2023gpt}                                           & 2023            & Not Available                     & Not Available                   \\ \hline
\multirow{3}{*}{EleutherAI}   & GPT-J\cite{wang2021gpt}                                           & 2021            & 6                                 & 825 GiB                         \\ \cline{2-5} 
                              & GPT-Neo\cite{gpt-neo}                                         & 2021            & 2.7                               & 825 GiB                         \\ \cline{2-5} 
                              & GPT-NeoX\cite{black2022gpt}                                       & 2022            & 20                                & 825 GiB                         \\ \hline
\multirow{4}{*}{DeepMind}     & Gopher\cite{rae2021scaling}                                          & 2021            & 280                               & 300 billion tokens              \\ \cline{2-5} 
                              & AlphaCode\cite{li2022competition}                                      & 2022            & 41                                & 967 billion tokens              \\ \cline{2-5} 
                              & Chinchilla\cite{hoffmann2022training}                                      & 2022            & 70                                & 1.4 trillion tokens             \\ \cline{2-5} 
                              & Sparrow\cite{glaese2022improving}                                         & 2022            & 70                                & Not Available                   \\ \hline
\multirow{4}{*}{Meta}         & Galactica\cite{taylor2022galactica}                                       & 2022            & 120                               & 106 billion tokens              \\ \cline{2-5} 
                              & OPT\cite{zhang2022opt}               & 2022            & 175                               & 180 billion tokens              \\ \cline{2-5} 
                              & OPT-IML\cite{iyer2022opt}                                         & 2022            & 175                               & Not Available                   \\ \cline{2-5} 
                              & LLaMA\cite{touvron2023llama}            & 2023            & 65                                & 1.4 trillion                    \\ \hline
\multirow{3}{*}{Hugging Face} & T0\cite{DBLP:journals/corr/abs-2110-08207}                                              & 2021            & 11                                & Not Available                   \\ \cline{2-5} 
                              & BLOOM\cite{workshop2022bloom}                                           & 2022            & 175                               & 350 billion tokens (1.6TB)      \\ \cline{2-5} 
                              & mT0\cite{muennighoff2022crosslingual}                                             & 2022            & 13                                & Not Available                   \\ \hline
\multirow{4}{*}{Baidu}        & Ernie 2.0 Large\cite{DBLP:journals/corr/abs-1907-12412}                                 & 2019            & 1.5                               & Not Available                   \\ \cline{2-5} 
                              & Ernie 3.0\cite{sun2021ernie}                                       & 2021            & 10                                & 375 billion tokens              \\ \cline{2-5} 
                              & Ernie 3.0 Titan\cite{wang2021ernie}                                 & 2021            & 260                               & 300 billion tokens              \\ \cline{2-5} 
                              & Ernie Bot (\url{https://yiyan.baidu.com/})                                      & 2023            & Not Available                     & Not Available                   \\ \hline
\end{tabular}
\end{table*}

\section{Background and Preliminaries}
For understanding the process of symbolic knowledge distillation of LLMs, we need to dive deeper into the two different technical theory of knowledge distillation followed by LLMs. Following sub-section will focus on that part.

\subsection{Knowledge Distillation}
Knowledge distillation is a technique used to transfer knowledge from a larger, more complex model (teacher) to a smaller, simpler model (student) with the goal of retaining much of the teacher model's performance\cite{gou2021knowledge}. This process is crucial in scenarios where computational resources are limited or where deployment requires lightweight models. There are various types of traditional knowledge distillation techniques: response-based, feature-based and relation-based and one modern symbolic knowledge distillation, each with its unique approach and area of application:

\subsubsection{Response-based Knowledge Distillation} 
Response-based knowledge distillation involves transferring knowledge from the teacher model's final output layer to the student model, aiming to mimic the teacher's final predictions. This approach is straightforward and has proven effective across various tasks, employing a loss function based on the divergence between the teacher's and student's logits. It's widely applied in model compression and has been adapted for different types of model predictions, including object detection and human pose estimation, where the teacher's output may include additional information like bounding box offsets\cite{chen2017learning} or heatmaps for landmarks\cite{zhang2019fast}. A key application of response-based knowledge distillation is in image classification\cite{hinton2015distilling}, where "soft targets" – the probabilities assigned to each class by the teacher model – play a crucial role. These probabilities are adjusted using a temperature factor to control the softness of the targets, allowing the transfer of knowledge from the teacher to the student. The distillation process typically employs the Kullback-Leibler divergence loss to optimize the similarity between the teacher's and student's probability distributions.

This method is praised for its simplicity and effectiveness, particularly in leveraging knowledge for training. However, its reliance on the final layer's output means it may not fully utilize intermediate-level supervision from the teacher, an aspect crucial for representation learning in deep neural networks.

\subsubsection{Feature-based Knowledge Distillation} Feature-based knowledge distillation taps into the strength of deep neural networks to learn hierarchical feature representations, a process central to representation learning\cite{bengio2013representation}. Unlike response-based knowledge distillation, which focuses on the outputs of the last layer, feature-based distillation utilizes the outputs from intermediate layers, or feature maps, to guide the student model. This approach is particularly beneficial for training models that are both narrower and deeper, as it provides a richer set of training signals.

The concept was first introduced with Fitnets\cite{romero2014fitnets}, aiming to improve student model training by matching feature activations between the teacher and student directly. Following this, several methodologies have been developed to facilitate this matching process, either directly or indirectly\cite{chen2021cross}. Notable contributions include the derivation of "attention maps" to express the use of neuron selectivity transfer\cite{huang2017like}, matching probability distributions in feature space\cite{passalis2018learning}, and introducing "factors" for more interpretable intermediate representations\cite{kim2018paraphrasing}. Techniques like route constrained hint learning\cite{jin2019knowledge} and the use of activation boundaries\cite{heo2019knowledge} have been proposed to minimize the performance gap between teacher and student models, alongside innovative strategies like cross-layer knowledge distillation\cite{chen2021cross} which adaptively matches teacher and student layers.

Despite the effectiveness of feature-based knowledge transfer in enriching the student model's learning, challenges remain in selecting appropriate layers for hints and guidance due to the size discrepancies between teacher and student models. This necessitates further exploration into how best to match the feature representations between teacher and student models effectively.

\subsubsection{Relation-based Knowledge Distillation} Relation-based knowledge distillation goes beyond the scope of response-based and feature-based methods by examining the relationships between different layers or data samples within the teacher model. This approach delves into the dynamics between feature maps, layers, and even the relationships between different teachers or data samples, offering a more nuanced form of knowledge transfer.

Flow of solution process (FSP)\cite{yim2017gift} utilizes the Gram matrix between two layers to encapsulate the relationships between pairs of feature maps through inner product calculations. Knowledge distillation via singular value decomposition\cite{lee2018self} distill essential information from these relationships. \cite{zhang2018better} explored multi-teacher scenarios by constructing graphs based on logits and features from each teacher, modeling their importance and relationships. \cite{lee2019graph} proposed a multi-head graph-based distillation technique that leverages intra-data relations between feature maps through a multi-head attention network. \cite{passalis2020heterogeneous} focused on pairwise hint information, allowing the student model to mimic mutual information flows from pairs of hint layers in the teacher model.

The distillation loss in relation-based knowledge distillation is formulated based on the similarity and correlation functions between the feature representations of teacher and student models, aiming to capture and transfer the intricate relationships present in the teacher's architecture. Relation-based knowledge can also encompass structured knowledge of data, privileged information about input features, and various other categories, each represented by different loss functions like Earth Mover distance, Huber loss, Angle-wise loss, and Frobenius norm. While recent advancements have introduced several types of relation-based knowledge, the challenge remains in effectively modeling the relational information from feature maps or data samples for knowledge transfer. This area continues to be ripe for further research and exploration to enhance the efficacy of knowledge distillation techniques.

\begin{figure}
\centerline{\includegraphics[width=18.5pc]{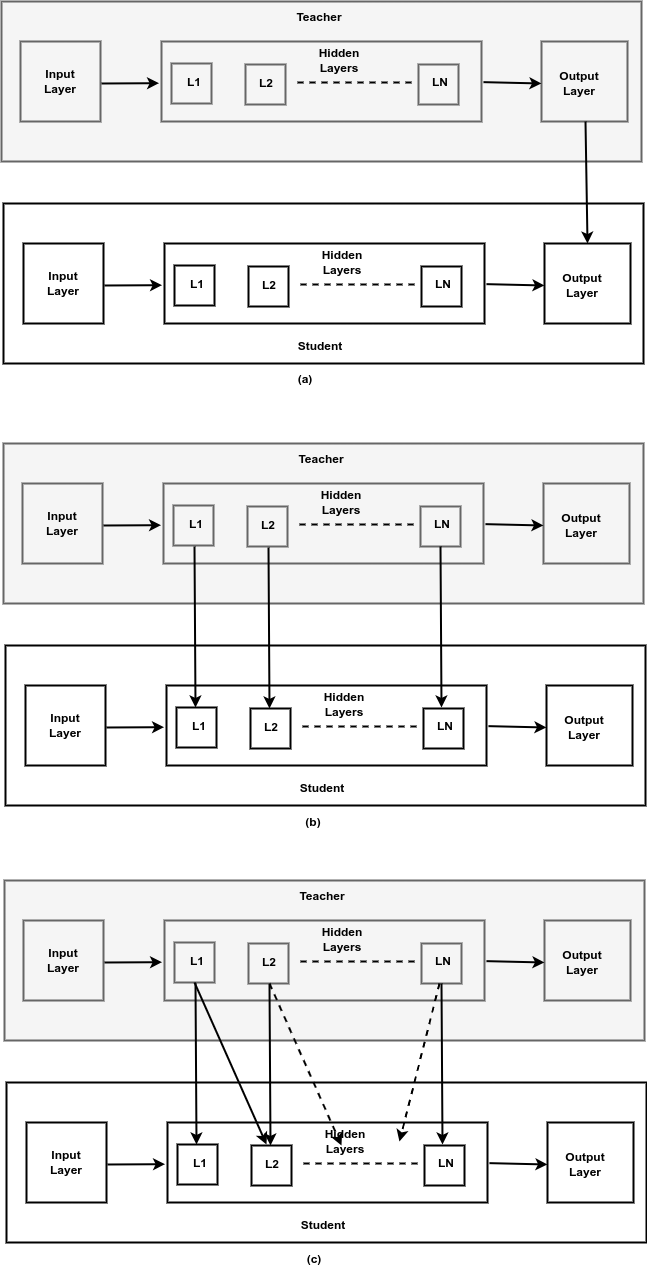}}
\caption{Types of Traditional Knowledge Distillation (a) Response-based, (b) Feature-based and (c) Relation-based}
\label{fig_tradKD}
\end{figure}

\subsubsection{Symbolic Knowledge Distillation}Contrary to the methods discussed earlier, symbolic knowledge distillation is centered on the distillation and transmission of knowledge in a symbolic format, including rules, logic, or symbolic representations. This method integrates structured knowledge bases and rules with machine learning models to boost their performance and clarity. It encodes intricate, structured information in a manner that allows for manipulation in reasoning, inference, and decision-making processes. The importance of this approach lies in its alignment with human methods of interpreting and reasoning with knowledge, thus providing enhanced transparency and interpretability.

Symbolic knowledge distillation represents a technique within machine learning where knowledge is extracted from a complex, typically less transparent model (like a deep neural network) and converted into a symbolic, more understandable format. This methodology merges the principles of conventional knowledge distillation with those of symbolic AI, aiming to improve the interpretability, transparency, and possibly the efficiency of machine learning models. It serves as a bridge between the often "black box" nature of deep learning models and the necessity for models that can be comprehended and trusted by humans. Such a requirement is especially critical in sectors demanding high levels of responsibility and explainability, including healthcare, finance, and autonomous driving. Although the specific mathematical model employed may vary based on the approach and the symbolic representation chosen, the overall process typically includes several defined steps.

\textbf{Training the Teacher Model:} A complex model (teacher) is trained on a dataset to achieve high performance. This model can be a deep neural network, and its architecture and training process depend on the specific task (e.g., image recognition, NLP).

\textbf{Extracting Knowledge:} 
The subsequent phase involves deriving insights from the teacher model, achievable through multiple approaches, including: examining the neuron activation patterns within the network; employing methods like Layer-wise Relevance Propagation (LRP)\cite{montavon2019layer} or  SHapley Additive exPlanations(SHAP)\cite{lundberg2017unified} to assess the significance of various inputs in the network's decision-making process; and identifying rules or patterns based on the decision boundaries established by the network.

\textbf{Symbolic Representation:} The gathered knowledge is subsequently converted into a symbolic representation. This process includes: developing decision trees or compiling sets of logical rules that mimic the neural network's behavior, and utilizing graphical models or alternative structured forms to encapsulate the relationships and dependencies deciphered by the network.

\textbf{Training the Student Model:} Following the translation of extracted knowledge into a symbolic form, a simpler and more interpretable 'student' model is trained to mimic this symbolic representation. The training process involves two key strategies. The symbolic representation may be used directly as a comprehensive set of rules for decision-making, allowing the student model to replicate decision processes based on predefined logical rules or the student model is trained to approximate the symbolic representation itself. This approach often incorporates conventional supervised learning techniques, with the significant distinction that the symbolic knowledge extracted from the teacher model acts as a guide or target.

\textbf{Evaluation and Refinement:} Once the student model has been trained to mimic the symbolic representation, it undergoes evaluation to verify that it retains the critical knowledge and performance attributes of the teacher model. This assessment might reveal the need for adjustments either to the symbolic representation itself or to the training methodology of the student model. Such refinements are crucial for ensuring that the student not only approximates the teacher's performance but does so in a way that is both interpretable and transparent. This emphasis on interpretability and transparency is key, as it aims to produce a student model that not only performs well but also provides insights into its decision-making processes, making it more understandable and trustworthy to users.

\begin{figure*}[htbp]
\centerline{\includegraphics[width=\textwidth]{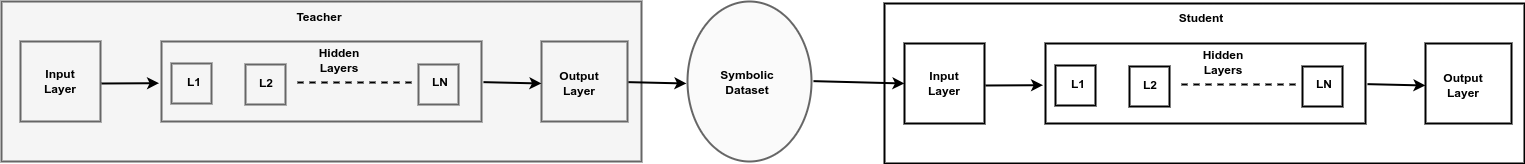}}
\caption{Symbolic Knowledge Distillation}
\label{fig_skd}
\end{figure*}

 \begin{table*}[]
 \caption{Comparison of Traditional and Symbolic Knowledge Distillation Process}
\centering
\begin{tabular}{|c|l|l|}
\hline
Parameters             & Traditional Knowledge Distillation      & Symbolic Knowledge Distillation          \\
\hline
Nature of Knowledge Transfer     & \makecell[l]{Soft outputs or logits which represent the teacher's \\ learned probability distribution}  & \makecell[l]{Human-readable representations such as logical rules,\\ decision trees, or graphical models}             \\
\hline
Interpretability and Transparency      & Student model remains a black-box neural network   &\makecell[l]{Student model, guided by symbolic representations offer \\ insights into the decision-making process}      \\
\hline
Methods Used for Distillation  & \makecell[l]{Techniques such as temperature scaling are used to \\ soften the teacher's outputs} & \makecell[l]{Involve methods like Layer-wise Relevance Propagation \\ (LRP) or SHAP }         \\
\hline
Student Model  & \makecell[l]{Mimic the teacher model} & \makecell[l]{Can be tune to behave differently than teacher model}         \\
\hline
Data Generation  & \makecell[l]{No} & \makecell[l]{Yes     }    \\
\hline
Layerwise Dependency  & \makecell[l]{Differnet layers have different influences} & \makecell[l]{No such dependency}         \\
\hline
\end{tabular}
\label{tab15}
\end{table*}

\subsection{Large Language Models}
LLMs are the foundation model for the language and has been the hot topic for past few years. Alot of opportunities has been created in one hand and due to ineffective use, it has also created some kind of fear among the users. In this section we will focus on the architecture of LLM followed by the training process.

\subsubsection{Architecture}
Transformer\cite{vaswani2017attention} architecture is the backbone of all the LLMs. Due to its features like parallelizable computation, attention based mechanism it has been able to reduced reliance in hand-crafted features and also improved the performance in NLP tasks. All the LLMs are directly or in-directly has the root the in the transformer architecture. Existing all the LLMs can be found to be belonging into one of the following architecture:

\textbf{Encoder-Decoder Architecture:} The underlying principle of this architecture involves transforming the input sequence into a fixed-length vector form, and subsequently, transforming this representation into the output sequence. The architecture is composed of two sets of Transformer blocks: one serving as the encoder and the other as the decoder. The encoder is tasked with processing the input sequence, utilizing a series of multi-head self-attention layers to convert it into latent representations. These representations are then leveraged by the decoder, which, through an autoregressive process, generates the output sequence by employing cross-attention mechanisms to focus on the latent representations provided by the encoder. PLM like T5\cite{raffel2020exploring}, BART\cite{lewis2019bart} and Flan-T5\cite{chung2022scaling} uses this architecture.

\textbf{Casual Decoder Architecture:} The causal decoder architecture is a type of decoder-only architecture used in language modeling, where the input and output tokens are processed in the same fashion through the decoder. This architecture incorporates a unidirectional attention mask, which ensures that each input token can only attend to past tokens and itself by masking all future attentions to zeros. The GPT-series models, including GPT-1\cite{radford2018improving}, GPT-2\cite{solaiman2019release}, and GPT-3\cite{brown2020language}, are representative language models of this architecture. Many other LLMs, such as OPT\cite{zhang2022opt}, BLOOM\cite{scao2022bloom}, and Gopher\cite{rae2021scaling}, have also adopted the causal decoder architecture.

\textbf{Prefix Decoder Architecture:} The prefix decoder architecture, also known as a non-causal decoder, is another type of decoder-only architecture which revises the masking mechanism of causal decoders to enable bidirectional attention over the prefix tokens, while maintaining unidirectional attention only on generated tokens. This allows the prefix decoders to bidirectionally encode the prefix sequence and predict the output tokens autoregressively, where the same parameters are shared during encoding and decoding. Unlike the causal decoder architecture, the prefix decoder architecture can incorporate bidirectional information into the decoding process, making it more suitable for tasks that require understanding the context of the entire input sequence. Existing representative LLMs based on prefix decoders include GLM-130B\cite{zeng2022glm} and U-PaLM\cite{tay2022transcending}.

\subsubsection{Training Process of Large Language Models}
The whole training process of LLM can be divided into two phases:

\textbf{Pre-trainning:}Pre-training LLMs involves training on extensive unlabeled text datasets to learn general language patterns and insights. The success of pre-training hinges on both the scale and quality of the training corpus, with large, diverse datasets allowing models to capture a wide array of language patterns and generalize effectively to new data.

The pre-training process unfolds in phases, starting with data collection, which is divided into general and specialized data sources. General data encompasses a wide range of text, including webpages, conversations, Q\&A portals, and books, while specialized data targets more niche content like research papers, code, and multilingual texts. The second phase, data pre-processing, focuses on refining the dataset by eliminating noisy, redundant, and irrelevant content. Techniques employed include quality filtering, deduplication (at sentence, document, and dataset levels), privacy protection (removing personal information), and tokenization (splitting text into manageable units for the model). Given that LLMs are not typically retrained frequently, the pre-training phase must be approached with precision, prioritizing a balanced mix of source materials\cite{rae2021scaling}, and ensuring both the quantity\cite{touvron2023llama} and quality\cite{hernandez2022scaling}  of the data are optimal. Pre-training tasks may involve language modeling\cite{radford2019language}, favored by decoder-only architectures for predicting subsequent tokens, or de-noising autoencoding\cite{lewis2019bart}, which focuses on correcting or replacing corrupted tokens.

\textbf{Fine tuning or Adaptive tuning:}
The fine-tuning stage is crucial for adapting pre-trained LLMs to specific domains or tasks, leveraging labeled examples or reinforcement learning to refine the model's understanding and predictive capabilities. It encompasses two main strategies: instruction tuning and alignment tuning.

Instruction tuning entails the fine-tuning of a language model by incorporating explicit instructions or demonstrations during training. This approach is designed to direct the model towards desired behaviors and outcomes, facilitating a more targeted response to tasks. The instructions for this tuning can be derived from existing datasets reformatted to include clear directives or crafted to reflect specific human needs. Alignment tuning, on the other hand, aims to adjust the LLM's outputs to match human expectations accurately, a process that may involve a trade-off known as the alignment tax\cite{glaese2022improving}. This concept refers to potential compromises in the model's capabilities as it is fine-tuned to prioritize outputs that are deemed more acceptable or beneficial from a human perspective. The most commonly used alignment criterias are helpfulness, honesty, and harmlessness\cite{glaese2022improving}\cite{ouyang2022training}. Few other criteria are also mentioned like behavior, intent, incentive, and inner aspects\cite{kenton2021alignment}.


\section{Symbolic Knowledge Distillation of Large Language Models}

Symbolic Knowledge Distillation of LLMs aimed at distilling the extensive knowledge encapsulated within LLMs into more interpretable and efficient forms. It's central methodology revolves around transforming the latent knowledge of models like GPT-3 into symbolic or rule-based representations. It involves a sophisticated process designed to transform the latent, complex knowledge within these models into explicit, structured, and interpretable forms. This process begins with the careful crafting of customised prompts that guide LLMs to generate outputs rich in specific knowledge types. Following this, NLP techniques like Named Entity Recognition (NER), Part-Of-Speech (POS) tagging, and dependency parsing, are employed to analyze and structure the responses. This step extract meaningful information and identify patterns within the text, which are then transformed into structured knowledge formats such as logic rules, knowledge graphs, or semantic frames. It derives explicit rules and patterns from the LLMs' responses, thereby facilitating the encoding of this information into symbolic representations that can be easily understood and manipulated.

The subsequent phase of this process involves the refinement and validation of the generated symbolic representations to preserve depth of knowledge and to ensure their accuracy, consistency, and practical utility. This includes refining the symbolic knowledge using the human experts or using the trained models to classify the generated knowledge on the basis of quality. The refined symbolic knowledge base undergoes validation against established benchmarks, allowing for the assessment of enhancements and ensuring the symbolic representations meet the required standards of quality and utility.

The creation of a high-quality knowledge base facilitates the training of smaller models, demonstrating that a quality dataset can significantly improve the performance of models that are 100 times smaller than their teacher counterparts\cite{west2021symbolic}. This highlights the efficacy of integrating symbolic knowledge into language models, presenting a viable alternative to scaling up LLMs. Symbolic knowledge distillation generates smaller, yet more efficient models, making them suitable for deployment in everyday practical applications, offering a more resource-efficient pathway to achieving high-quality outputs in language models.

Various approaches that are used to distill the symbolic knowledge of LLMs can be categorised as:
\subsection{Direct Distillation}
The distillation of symbolic knowledge from LLMs like GPT-3 begins with the construction of a specific prompt. This prompt is designed to elicit responses that encapsulate commonsense or factual understanding. It could involve scenarios, questions, or statements that require the application of general knowledge about the world. The effectiveness of this step hinges on the ability to craft prompts that are both clear and contextually rich enough to guide the LLM towards producing relevant and insightful outputs. Upon receiving the prompt, the LLM generates a response based on its training and the intricacies of the provided context. These models, have been exposed to extensive and varied textual data, encompassing a wide array of commonsense situations and factual knowledge. This extensive training enables them to generate responses that are not only contextually appropriate but also rich in commonsense and factual knowledge. The model's response is a complex interplay of its learned patterns, linguistic understanding, and the implicit knowledge embedded within its training corpus. This step translates the implicit knowledge within the model into explicit textual responses that can be further analyzed and utilized for knowledge extraction. 

The generated text is then analyzed to extract knowledge. This can be in the form of statements, inferences, or relationships that are implicitly or explicitly expressed in the text. The extraction process might involve additional processing steps like parsing the text to identify relevant information or using templates to format the knowledge in a structured way. The knowledge base derived from this process can be further improved with the assistance of critics, who may be human evaluators providing feedback on the quality and acceptability of the generated content. Once a substantial volume of high-quality generated data has been accumulated, this data can be utilized to train a critic model like RoBERTa, which can be used to evaluate the generated text for accuracy, relevance, and coherence. The critic model can filter out lower-quality outputs, ensuring that only high-quality commonsense knowledge is retained. The high-quality knowledge can then be distilled into structured formats like knowledge graphs or further trained into specialized models. This process involves organizing the knowledge in a way that can be easily utilized by other systems or models.
\begin{figure}
\centerline{\includegraphics[width=18.5pc]{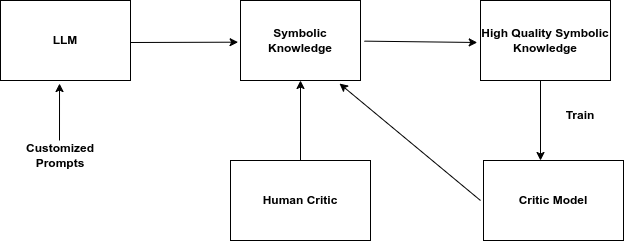}}
\caption{Overview of Direct Distillation process LLMs}
\label{fig_23}
\end{figure}

\subsection{Multilevel distillation of symbolic knowledge}

This approach iteratively refines the knowledge transfer from a larger, pre-trained teacher model to a smaller, more efficient student model. The process begins with the teacher model, typically a LLM like GPT-3, generating initial knowledge base. The generated knowledge base is then filtered for quality, focusing on aspects like accuracy and length. The smaller student model, such as GPT2-Large, is initially trained on this filtered dataset. Subsequently, the student model generates new knowledge base, which are again filtered to enhance quality. This cycle of generation and refining through filtering is repeated iteratively, with each iteration aiming to improve fidelity and succinctness of the distilled knowledge.

During each iteration, various filters are applied to ensure the quality which are fidelity filter, length filter or contextual filter. The Fidelity Filter ensures a true representation of the input sentence, verified using an off-the-shelf Natural Language Inference (NLI) model. The Length Filter controls the length to fit within a predefined compression ratio, gradually guiding the model to produce increasingly concise output. A Contextual Filter is used in some cases, focusing on the coherence in the larger context of the text. The process results in the development of increasingly efficient student models that inherit the distillation ability of the teacher model but with enhanced control over quality. This method allows for the creation of high-quality, succinct dataset with diverse compression ratios, without relying on pre-existing annotated datasets.
\begin{figure}
\centerline{\includegraphics[width=18.5pc]{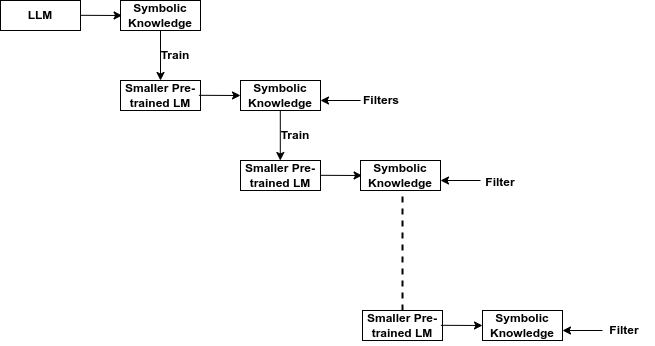}}
\caption{Overview of Multilevel Distillation process LLMs}
\label{fig_232}
\end{figure}

\subsection{Distillation using Reinforcement Learning policy}

The approach refines the policy of a LLM through a two-step iterative process: generating and filtering data. The first step, involves using the current LLM policy to generate a range of output predictions for given contexts, effectively augmenting the training dataset. Initially, this policy might be based on a supervised learning model, and the generated outputs may not be perfectly aligned with human preferences. However, this step is essential for creating a diverse set of potential outputs for further refinement. The generated data forms the basis for the next critical phase of the process.

In the second step, the data produced is ranked and filtered using a filters like scoring function, typically a learned reward model trained on human preferences. This step is pivotal in selecting the best outputs that align with the desired human outcomes, as determined by the scores from the reward model. The filtering threshold can be incrementally increased in subsequent iterations, ensuring that only the top-performing outputs are selected for further training. The language model is then fine-tuned on this curated dataset with an offline RL objective, adjusting its policy to produce outputs that are more likely to receive high scores. This process of generating and filtering, repeated iteratively, serves as a feedback loop, continuously refining the model's policy towards outputs increasingly aligned with human preferences. 

All three techniques mentioned have been successfully applied to various research areas, including commonsense reasoning\cite{west2021symbolic}, translation\cite{gulcehre2023reinforced}, summarisation\cite{sclar2022referee} , and mathematical reasoning\cite{polu2022formal}, among others, yielding significant results. $\hyperref[fig_RW]{Fig. 7}$ provides an overview of all the areas explored so far, with detailed discussions presented in the related works section. $\hyperref[tab14]{Table. III}$ offers insights into each research area, categorizing them based on the techniques discussed above.

\begin{figure}
\centerline{\includegraphics[width=18.5pc]{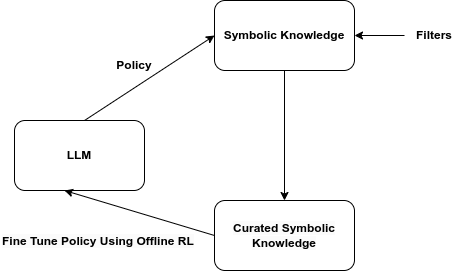}}
\caption{Overview of Distillation process using RL}
\label{fig_2321}
\end{figure}
\section{Related Works}
\begin{figure*}[htbp]
\centerline{\includegraphics[width=\textwidth]{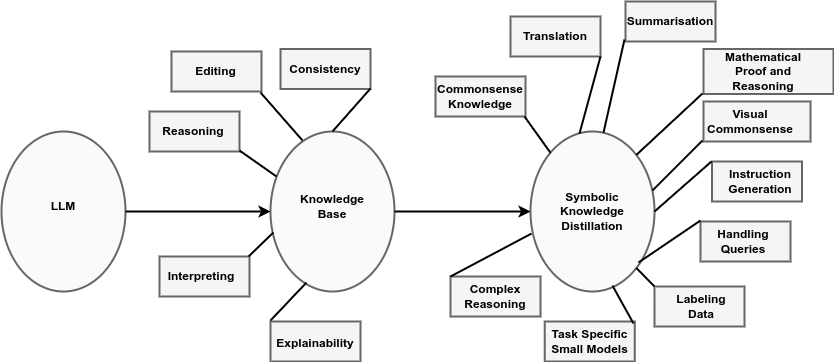}}
\caption{Overview of Related Works}
\label{fig_RW}
\end{figure*}

In this segment, we begin by exploring the foundational work that positions LLMs as a knowledge base and then delve into research focused on analyzing the knowledge contained within LLMs. Lastly, we review efforts aimed at distilling this knowledge into a symbolic form. An overview of this concept is presented in
$\hyperref[fig_RW]{Fig. 7}$.
\subsection{Knowledge Base of LLM}

LLM can act as a knowledge base or oracle that performs well on open-domain question answering without fine-tuning\cite{petroni2019language}. LLM can also function as the domain-specific KBs in biomedical field however they are highly influenced by prompt bias and synonym variance\cite{sung2021can}. It rapidly and stably acquires linguistic knowledge, including syntax, grammar, and parts of speech, predominantly in the early stages of pre-training, showing little variation across different domains. In contrast, the assimilation of factual and commonsense knowledge is slower, more sensitive to the domain of the training data, and exhibits a more gradual progression throughout the pre-training period\cite{liu2021probing}.

\subsection{Consistency of Knowledge In LM}
The research\cite{elazar2021measuring} sheds light on the consistency of knowledge in PLMs like BERT and RoBERTa. Their findings reveal a concerning lack of consistency in these models, particularly when responding to paraphrased queries with factual content. The study\cite{kassner-schutze-2020-negated} adds another layer of complexity to this issue by highlighting the challenges PLMs face in accurately processing negated facts and their susceptibility to being misled by contextually irrelevant or misleading information.

\subsection{Editing the Knowledge in LLM}
Editing knowledge in LLMs has become a prominent area of research with several innovative approaches proposed to address this challenge. Constrained layer-wise fine-tuning\cite{zhu2020modifying} formulates knowledge modification as a constrained optimization problem and allows for fine-tuning specific layers to update knowledge while retaining existing information. \cite{dai2021knowledge} introduced the concept of Knowledge Neurons, enabling pinpointing specific components responsible for factual knowledge within LLMs and providing the means to manipulate them for altering model output. The KNOWLEDGEEDITOR\cite{de2021editing} offers an efficient way to update factual knowledge in pre-trained LLMs without extensive retraining.The paper\cite{hase2021language} introduces methods for detecting, updating, and visualizing beliefs in LLM by using the Sequential Local and Generalizing (SLAG) update objective.  Model Editor Networks with Gradient Decomposition (MEND)\cite{mitchell2021fast} efficiently edit large-scale pre-trained models by transforming gradients during fine-tuning. Continual Knowledge Learning (CKL)\cite{jang2021towards} addresses the challenge of updating and maintaining the relevancy of world knowledge in LLMs. 

\subsection{Reasoning with Knowledge in LLM}
The research landscape concerning reasoning abilities in PLMs and transformers, has seen significant exploration and development.
The paper\cite{kassner2020pretrained} found that while BERT could learn simpler one-hop rules, it struggled with more complex two-hop rules and distinguishing between symmetric and non-symmetric relations. \cite{clark2020transformers} demonstrates that transformers can effectively emulate reasoning over language, achieving high accuracy on various synthetic datasets that require different depths of inference and can act as limited "soft theorem provers". PROVER\cite{saha2020prover} extended \cite{clark2020transformers} to answer binary questions over rule-bases while generating corresponding proofs for enhanced interpretability. ProofWriter\cite{tafjord2020proofwriter} stands out for its ability to produce implications and corresponding natural language proofs from given theories, using the T5 transformer architecture. The paper\cite{gontier2020measuring} explores the capability of Transformer Language Models (TLMs) in logical reasoning with natural language focusing on first-order logic proofs. The paper\cite{richardson2022pushing} explore the capacity of transformer models to perform deductive reasoning on logical theories expressed in natural language by introducing a method for generating challenging reasoning datasets whereas the paper\cite{saeed2021rulebert} enhance the deductive reasoning abilities of PLMs using soft Horn rules and achieved high performance on unseen logical rules and showed improved understanding of logical properties like negation and symmetry. The paper\cite{saxton2019analysing} introduces a novel dataset to evaluate the mathematical reasoning capabilities of neural networks, focusing on problems across arithmetic, algebra, probability, and calculus.

The paper\cite{banerjee2021commonsense} integrates commonsense reasoning on natural language question-answering tasks by employing smaller language models,and demonstrate competitive performance against large PLMs. RICA (Robust Inference using Commonsense Axioms)\cite{zhou2020rica},found that PLMs are vulnerable to perturbation attacks, where minor changes in input data drastically alter their conclusions. The paper\cite{rajani2019explain} presents the Common Sense Explanations (CoS-E) dataset and the Commonsense Auto-Generated Explanation (CAGE) framework, which leverages natural language explanations(human-like explanations) to improve model's reasoning capabilities.

\subsection{Interpreting the Knowledge of LLM}
Interpreting the knowledge encoded in LLMs has been advanced through various studies, each contributing unique insights into how these models capture and process linguistic information.
\cite{jain2019attention} argue that attention weights often don't align with other feature importance measures and can produce similar predictions despite different attention distributions. This view is nuanced by \cite{wiegreffe2019attention}, who suggest that attention can serve as an explanation, but its validity depends on the context and testing methods. \cite{serrano2019attention} also investigate attention in text classification, finding that while there is some correlation between attention weights and model predictions, attention weights alone are not definitive indicators of input importance and propose that gradient-based attention weight rankings provide a deeper understanding.

The study\cite{zhao2021non} include method for quantifying non-linearity in transformers, particularly in feed-forward networks. They reveal a non-distinct feature extraction process in BERT layers, influenced by skip connections. \cite{geva2020transformer} demonstrate that transformer layers function as key-value memories, capturing textual patterns and inducing distributions over the output vocabulary, with lower layers focusing on shallow patterns and upper layers on semantic ones. \cite{meng2022locating} show that factual associations in GPT models are tied to localized computations, particularly in middle-layer feed-forward modules.

\subsection{Explainability in LLM}
 The study\cite{han2020explaining} investigates the application of Influence Functions (IFs) to identify artifacts in models, comparing their effectiveness with that of common word-saliency methods. Researchers in study \cite{pezeshkpour2021empirical} compare IFs with simpler retrieval-based methods and suggest that despite the complexity of IFs, simpler methods can achieve comparable performance. Exploring further in study\cite{pezeshkpour2021combining}, they introduce Training-feature attribution (TFA), which synergizes saliency maps and instance attribution to effectively uncover artifacts. Researcher in \cite{zylberajch2021hildif} propose Human In the Loop Debugging using Influence Functions (HILDIF), a pipeline that employs influence functions for debugging deep text classifiers, allowing human involvement in enhancing model performance.

In a different approach, study \cite{narang2020wt5} presents a novel method for training language models to generate natural text explanations alongside their predictions, utilizing the text-to-text framework\cite{raffel2020exploring}. Addressing the challenge of inconsistency in natural language explanations, \cite{camburu2019make} introduces an adversarial framework to identify and measure these inconsistencies. The Proto-Trex model\cite{friedrich2021interactively} uses prototypical examples to explain model predictions, thus mitigating the opacity often associated with complex models. Research\cite{lei2016rationalizing} enhances interpretability by extracting key text segments, termed "rationales", serving as justifications for model predictions. Study\cite{paranjape2021prompting} works on improving commonsense reasoning by employing contrastive explanations generated through specialized prompts, aligning model reasoning more closely with human cognitive patterns.

\subsection{Symbolic Knowledge Distillation} The  conducted research works in this area can be categorised as follows:
\subsubsection{Commonsense Knowledge}
The study\cite{west2021symbolic} introduces a transformative shift in the conventional practice, transitioning from the traditional 'from-human-to-corpus-to-machine' approach to an innovative 'from-machine-to-corpus-to-machine' paradigm through the introduction of symbolic knowledge distillation. In their research, the authors not only succeed in creating a substantially larger common-sense dataset from ATOMIC resource\cite{sap2019atomic}, approximately ten times larger than previously manually synthesized datasets, but also enhance its diversity and quality. Their novel approach involves training the common-sense model using this newly generated knowledge graph. Despite being only 1/100th of its predecessor model, it outperforms the previous model, showcasing the effectiveness of their approach. The paper\cite{west2023novacomet} introduces NOVACOMET, an innovative open commonsense knowledge model that merges the strengths of both knowledge and general task models. This model, built upon symbolic knowledge distilled from proprietary models like GPT-3, creates an auditable discrete knowledge graph, NOVATOMIC, which facilitates open-format training and application to a wide array of reasoning tasks. It demonstrates superior performance in commonsense reasoning, outperforming comparable models in various benchmarks. The model's training involves novel techniques like commonsense field masking for enhanced flexibility in knowledge handling. Iterative Imitation and Decoding for Distillation(I2D2)\cite{bhagavatula2022i2d2} framework employs a four-stage process that includes prompt construction, constrained decoding using NeuroLogic Decoding, critic filtering, and self-imitation learning, where the model is iteratively refined based on its own high-quality outputs. A new corpus, Gen-A-tomic, was created to provide diverse and accurate commonsense knowledge. I2D2 demonstrated superior performance in accuracy and precision over larger models like GPT-3, with GPT-2 XL showing significant improvements through self-imitation learning iterations.
\subsubsection{Translation}
Reinforced Self-Training (ReST)\cite{gulcehre2023reinforced}  is a method to align LLMs with human preferences in the realm of machine translation. This approach incorporates reinforcement learning from human feedback (RLHF) to enhance the output quality. ReST initiates by generating a dataset through sampling from the initial LLM policy, followed by the application of offline reinforcement learning algorithms to refine the policy. This method is identified as more efficient than traditional online RLHF techniques, primarily because it facilitates the creation of the training dataset in an offline manner, promoting the reuse of data. The effectiveness of ReST is demonstrated through significant improvements in translation quality, validated by both automated metrics and human evaluations across various machine translation benchmarks.
\subsubsection{Summarisation}
REFEREE\cite{sclar2022referee} is a framework for reference-free sentence summarization that allows for direct control of compression ratio. It uses Symbolic Knowledge Distillation to distill latent knowledge from PLMs, resulting in smaller but better summarizers with sharper controllability. The framework employs iterative distillation of knowledge, where student models from previous iterations serve as teacher models in the next iteration. This iterative process also generates a high-quality dataset of sentence-summary pairs with varying compression ratios. The final student models outperform the larger GPT3-Instruct model in terms of compression ratio controllability without compromising the quality of the summarization.
\subsubsection{Mathematical Proof and Reasoning}
The paper\cite{polu2022formal} presents a method called expert iteration, which combines proof search with learning to improve language modeling in formal mathematics. The method involves finding new original proofs for the same statements and closing marginally harder statements at each iteration, which in turn provides more useful training data for the next iteration. By interleaving proof search with learning, expert iteration is able to dramatically outperform proof search only. The paper demonstrates the effectiveness of expert iteration on a manually curated set of problem statements and achieves state-of-the-art results on the miniF2F benchmark, a set of formalized statements of mathematical problems from various competitions.The paper\cite{fu2023specializing} explores the concept of distilling abilities from LLMs into smaller ones, specifically for enhancing their performance in multi-step math reasoning tasks. The process begins with generating a dataset using a larger model (like GPT-3.5) employing chain-of-thought reasoning, where the model details the steps leading to a solution. This dataset is then used to fine-tune a smaller T5 model, with the aim of specializing its abilities in the specific area of multi-step reasoning. This fine-tuning process allows the smaller model to learn the complex reasoning patterns demonstrated by the larger model.

\subsubsection{Visual Commonsense}
Localized Symbolic Knowledge Distillation (LSKD)\cite{park2023localized} enhances vision-language models by focusing on localized regions within images. This method addresses a significant limitation in existing models, which interpret images as a whole, by introducing Localized Visual Commonsense models that can specify and reason about multiple distinct regions in an image. The authors develop a scalable framework for generating localized visual commonsense statements and establish the Localized Commonsense Knowledge Corpus, which aids in expanding the capabilities of vision+language models to include references-as-input. The paper highlights the state-of-the-art zero-shot performance of these models on three localized visual reasoning tasks and showcases the superiority of the student model over the teacher model through human evaluation.

\subsubsection{Instruction Generation}
Traditional instruction-tuned models, reliant on human-written instruction data, often lack diversity and creativity, constraining the generality of the model. SELF-INSTRUCT\cite{wang2022self}  mitigates this by enabling models to generate their own instructions, inputs, and outputs, which are then used for fine-tuning. This process involves generating task instructions, classifying them, creating instances via input-first or output-first approaches, and filtering out low-quality data. The approach significantly reduces the need for human-labeled data, fostering a broader and more creative instructional capability in LMs. The performance evaluation shows that the GPT3SELF-INST model, fine-tuned on this self-generated data, substantially outperforms the vanilla GPT-3 in instruction-following tasks and closely matches the performance of models like InstructGPT001. Alpaca\cite{alpaca} enhance the SELF-INSTRUCT data generation pipeline by employing the more advanced text-davinci-003 model for instruction data generation that explicitly defines the requirements for instruction generation, aiming for more focused and relevant outputs. The adoption of aggressive batch decoding, producing 20 instructions simultaneously, significantly reduces data generation costs and simplifying the pipeline by eliminating the distinction between classification and non-classification instructions and generating only a single instance per instruction, instead of 2 to 3, streamlines the process. Evol-Instruct\cite{xu2023wizardlm} is a novel method that uses LLMs to automatically generate a vast array of complex instructional data. This approach begins with simple initial instructions and employs the LLM to evolve these into more sophisticated and diverse instructions through in-depth and in-breadth evolution processes. It enhances instructions by adding constraints, increasing reasoning complexity, and diversifying topics, thus creating a rich dataset for fine-tuning LLMs. This dataset is used to train the LLaMA model, resulting in WizardLM, a model demonstrating superior performance in following complex instructions compared to human-generated datasets and existing models like ChatGPT.

\subsubsection{Handling queries}
Vicuna-13B\cite{vicuna2023} is an open-source chatbot developed by fine-tuning the LLaMA model with around 70,000 user-shared ChatGPT conversations from ShareGPT. It demonstrates superior performance, achieving over 90\% of ChatGPT's quality, and surpassing other models like LLaMA and Stanford Alpaca. The training, which cost approximately \$300, utilized advanced techniques for handling multi-turn conversations. Despite its advancements, Vicuna-13B shares common LLM limitations, such as challenges in reasoning or math tasks, and has potential issues with factual accuracy and safety. Koala\cite{koala_blogpost_2023}, a chatbot model developed by fine-tuning Meta’s LLaMA with web-sourced dialogue data, including interactions with large models like ChatGPT. Koala demonstrates competitive performance against established models such as ChatGPT and Stanford’s Alpaca, particularly in handling real user queries. ASK ME ANYTHING PROMPTING (AMA)\cite{arora2022ask} is a prompting method for improving the performance of LLMs like GPT-3. AMA leverages multiple effective but imperfect prompts, aggregating them using weak supervision to enhance prediction quality. This method primarily utilizes open-ended question-answering formats, which are found to be more effective than restrictive prompts. AMA's recursive use of the LLM to transform task inputs into these formats, combined with the aggregation of diverse prompts, demonstrates significant improvements in LLM predictions. QAMELEON\cite{agrawal2022qameleon} is an innovative approach to multilingual question answering (QA) systems, leveraging PLMs within a few-shot learning framework. PLMs generate QA pairs in multiple languages, significantly reducing the need for extensive, language-specific training datasets. By requiring only a minimal number of examples (as few as five per language), QAMELEON efficiently fine-tunes QA models, overcoming traditional constraints of resource-intensive data annotation. This approach not only simplifies and accelerates the development of multilingual QA systems but also achieves superior accuracy and efficiency, demonstrating its potential as a scalable and effective solution in NLP.
 
\subsubsection{Labeling Data}
The research paper\cite{wang2021want} examines the efficacy of using GPT-3 for data labeling in NLP tasks, highlighting its cost-effectiveness compared to traditional human labeling. The study reveals that GPT-3 can reduce labeling costs by 50\% to 96\% across various tasks, including sentiment analysis, text classification, and summarization. The paper introduces a novel framework that combines GPT-3 generated pseudo labels with human labels, improving performance under limited budgets. Furthermore, an active labeling strategy is explored, where low-confidence labels by GPT-3 are re-annotated by humans, enhancing label quality. Despite these benefits, the paper notes that GPT-3 is more suited for low-stakes labeling tasks, as its reliability in high-stakes scenarios remains limited. The research\cite{smith2022language} presents a novel method for utilizing PLMs in tasks with scarce labeled training data. This technique involves prompting the LM with multiple queries about an example, and the model's responses are then interpreted as votes for specific labels or as abstentions. This process, integrated within a weak supervision framework, leverages the capabilities of the LM as a labeling function. The Snorkel system is subsequently employed to clean and refine these noisy label sources, culminating in the creation of enhanced training data for an end classifier.

\subsubsection{Task Specific Small Models}
 The method, "Distilling step-by-step"\cite{hsieh2023distilling}, involves extracting rationales from LLMs alongside output labels. These rationales, serving as detailed explanations for model predictions, are then used in a multi-task learning framework to train smaller models on both label and rationale prediction tasks. This technique significantly reduces the data and model size required, enabling smaller models to surpass the performance of LLMs more efficiently. The paper demonstrates the effectiveness of this approach across multiple datasets and tasks, showcasing it as a resource-efficient alternative to standard finetuning and traditional distillation methods.
 
\subsubsection{Complex Reasoning}
 Orca \cite{mukherjee2023orca} is designed to enhance the capabilities of smaller models through imitation learning from large foundation models (LFMs). Traditional methods faced issues like limited imitation signals, small-scale homogeneous training data, and inadequate evaluation, leading to an overestimation of the small models' capabilities. These models often imitated the style but not the reasoning process of LFMs. Orca addresses these challenges by learning from GPT-4's rich signals, including explanation traces, step-by-step thought processes, and complex instructions, with guidance from ChatGPT as a teacher. This approach enables progressive learning through large-scale and diverse imitation data. Orca significantly outperforms state-of-the-art instruction-tuned models like Vicuna-13B in complex zero-shot reasoning benchmarks, achieving more than a 100\% improvement in Big-Bench Hard (BBH) and a 42\% improvement in AGIEval. Orca reaches parity with ChatGPT in BBH and exhibits competitive performance in professional and academic exams like the SAT, LSAT, GRE, and GMAT, in zero-shot settings without Chain of Thought (CoT), though it still trails behind GPT-4. Orca 2\cite{mitra2023orca} builds upon the Orca project, focusing on enhancing smaller LMs' reasoning abilities. Orca 2 continues exploration, particularly addressing the limitations of imitation learning, which had been the primary method for training small LMs. This method, while effective in replicating the output of larger models, often fell short in reasoning and comprehension skills. It introduces various reasoning techniques (e.g., step-by-step processing, recall-then-generate, recall-reason-generate, extract-generate, direct-answer methods) and focuses on teaching small LMs to choose the most effective reasoning strategy for a given task. This approach aims to enable small LMs to perform at their best, regardless of their size, by utilizing more nuanced data and training strategies. The system is described as a "Cautious Reasoner," learning to execute specific reasoning steps and strategize at a higher level how to approach particular tasks.
  \begin{table}
\caption{Related Works in Symbolic Knowledge Distillation}
\centering
\begin{tabular}{|c|c|c|}
\hline
       Research      & Types      & Application              \\
\hline
\cite{west2021symbolic}      & Direct  &Commonsense Reasoning             \\
\hline
\cite{sclar2022referee}       & Multi-level   &Summarisation   \\
\hline
\cite{gulcehre2023reinforced}  & RL based &Translation         \\
\hline
\cite{west2023novacomet}  & Direct  &Commonsense Reasoning    \\
\hline
\cite{bhagavatula2022i2d2}  & Direct &Commonsense Reasoning \\
\hline
\cite{polu2022formal}     & Direct  &Mathematical Proof and Reasoning           \\
\hline
\cite{fu2023specializing}     & Direct    &Mathematical Proof and Reasoning \\
\hline
\cite{park2023localized}     & Direct     &Visual Commonsense Reasoning  \\
\hline
\cite{wang2022self}  & Direct             &Instruction Generation \\
\hline

\cite{alpaca}  &  Direct  &Instruction Generation        \\
\hline
\cite{xu2023wizardlm}      & Direct &Instruction Generation \\
\hline
\cite{vicuna2023}  & Direct   &Handling Queries      \\
\hline
\cite{koala_blogpost_2023}       &   Direct   &  Handling Queries       \\
\hline
\cite{arora2022ask}        &   Direct  &  Handling Queries      \\
\hline
\cite{agrawal2022qameleon}     &   Direct &  Handling Queries \\
\hline
\cite{wang2021want} &   Direct    &  Labeling Data       \\
\hline
\cite{smith2022language} &     Direct  &  Labeling Data    \\
\hline
\cite{hsieh2023distilling}  &   Direct &  Generating Task Specific Small Models  \\
\hline
\cite{mukherjee2023orca}      &   Direct       &  Complex Reasoning \\
\hline
\cite{mitra2023orca}         &   Direct      &  Complex Reasoning    \\
\hline
\end{tabular}
\label{tab14}
\end{table}

 \begin{table*}
\caption{  Related Works in Symbolic Knowledge Distillation with their major components}
\centering
\begin{tabular}{|c|c|c|c|c|}
\hline
         Research      &   Teacher      &   Student &   Dataset Generated &   Size of Dataset              \\
\hline
\cite{west2021symbolic}      &   GPT-3(175B)  &   $COMET^{distil}(1.5B)$ &   Commonsense Knowledge Graph &   6.5M           \\
\hline
\cite{sclar2022referee}       &   GPT-3  &  REFEREE-CONTROL &  Sentence-summary pairs   &   100K    \\
\hline
\cite{gulcehre2023reinforced}  &   Encoder-Decoder Architecture &   Teacher Itself   &  Translation Dataset    &  N/A        \\
\hline
\cite{west2023novacomet}  &   GPT-3  &  NOVACOMET  &  NOVATOMIC   &   2.2M   \\
\hline
\cite{bhagavatula2022i2d2}  &   GPT-3 &  GPT-2  &   Gen-A-tomic   &  7M\\
\hline
\cite{polu2022formal}     &   Decoder Only Architecture  &  Teacher Itself   &   Tactic Dataset    &  N/A             \\
\hline
\cite{fu2023specializing}     &   GPT-3.5    &  FlanT5 &  Math Reasoning   &   N/A \\
\hline
\cite{park2023localized}     &   ChatGPT     &  BLIP-2  &   Localized Commonsense Knowledge    &  1M  \\
\hline
\cite{wang2022self}  &   GPT-3             &  Teacher Itself&  Instruction Dataset &   82K\\
\hline

\cite{alpaca}  &     GPT-3.5  &  7B LLaMA   &  Instruction Dataset &  52K       \\
\hline
\cite{xu2023wizardlm}      &   ChatGPT &  WizardLM &  Instruction Dataset  &  250K \\
\hline
\cite{vicuna2023}  &   ChatGPT   &  Vicuna-13B &  Conversational Dataset &   70K        \\
\hline
\cite{koala_blogpost_2023}       &   ChatGPT  &  Koala-13B  &  Conversational Dataset &   N/A      \\
\hline
\cite{arora2022ask}        &   GPT3-175B  &  GPT-J-6B   &  Prompt Dataset &  N/A      \\
\hline
\cite{agrawal2022qameleon}     &   PaLM-540B &   mT5-XL &  Multilingual QA    &  47173 \\
\hline
\cite{wang2021want} &    GPT-3    &  RoBERTa &  Labeled Data &   5.1K       \\
\hline
\cite{smith2022language} &     GPT-3  &  T0++ &  Labeled Data  &      N/A     \\
\hline
\cite{hsieh2023distilling}  &   540B PaLM &  770M T5  &  Rationales &  N/A \\
\hline
\cite{mukherjee2023orca}      &   GPT-4       &  Orca(13B)  &  Zero shot queries    &  5M \\
\hline
\cite{mitra2023orca}         &   GPT-4      &  Orca-2  &    Progressive queries &    817K      \\
\hline
\end{tabular}
\label{tab4}
\end{table*}

\section{Opportunities}
Symbolic Knowledge distillation of LLM has been one of the heated topics and has been gaining rapid popularity. Among the various areas, the most prominent areas where it can be applied are:
\subsection{Creation of larger, diversified and qualitative dataset}
It offers significant potential in enhancing dataset quality and diversity. This process involves extracting structured knowledge from LLMs to create datasets that are not only larger in scale but also exhibit a broader range of qualities and characteristics. These enriched datasets can be pivotal in training more robust and efficient machine learning models, leading to advancements in various domains such as NLP, image recognition, and beyond. The ability to generate high-quality datasets from LLMs accelerates the development of more sophisticated AI systems, contributing to advances in both academic research and practical applications.
\subsection{Reduction in the cost by utilising machines in the low level task under guidance on humans}Implementing symbolic knowledge distillation in low-level tasks allows for the effective delegation of routine and repetitive tasks to machines, significantly reducing operational costs. By leveraging the distilled knowledge from LLMs, machines can perform these tasks with a high degree of accuracy and efficiency, under the supervision of human experts. This collaboration between human intelligence and machine capabilities leads to optimized resource utilization, where humans focus on more complex, creative, or decision-making tasks while machines handle the routine aspects, thereby enhancing overall productivity and cost-effectiveness.
\subsection{Smaller and more powerful models than LLMs for summarization, translation, common sense etc} Distilling knowledge from LLMs into smaller models presents a promising avenue for creating compact yet powerful AI tools. These distilled models retain the core capabilities of their larger counterparts but with reduced computational requirements. This makes them particularly suitable for applications like text summarization, language translation, and common sense reasoning, where efficiency and speed are crucial. These smaller models offer the dual benefits of lower resource consumption and faster processing times, making them ideal for deployment in environments with limited computational resources or for applications requiring real-time responses.
\subsection{ Instruction tuning}
Instruction tuning, in the context of symbolic knowledge distillation from LLMs, refers to the process of refining and optimizing AI models to better understand and execute specific instructions. This approach enhances the model's ability to interpret and act upon user commands accurately, leading to more intuitive and user-friendly AI systems. Instruction tuning is particularly relevant in applications where user interaction is key, such as virtual assistants, educational tools, and interactive AI systems. By focusing on instruction tuning, developers can create AI models that are not only powerful in their capabilities but also align closely with user expectations and needs, facilitating more effective and seamless human-AI interactions.

\subsection{  Novel Algorithm and Evaluation Benchmark}
  Size alone does not determine the quality of language generation. Innovative approaches, such as those seen in I2D2\cite{bhagavatula2022i2d2}, present a viable option, particularly in scenarios where utilizing massive models like GPT-3 is impractical. Given that this field is in its infancy, the evaluation benchmarks are quite intricate and require significant refinement. Current evaluation techniques are from traditional knowledge distillation benchmarks and must be updated to fit this novel area of study.Symbolic Knowledge Distillation of LLMs involves two components: the neural aspect (LLMs) and the symbolic aspect (distilled symbolic knowledge). Together, these form a Neurosymbolic model, which necessitates the development of new benchmarks for evaluation, testing, and validation\cite{renkhoff2024survey}.

\subsection{Creation of Open source data and open model}
The concept of symbolic distillation presents an intriguing avenue for creating open source data and models within the realm of LLMs. Currently, many LLMs are proprietary and trained on closed-source data, limiting accessibility and transparency. Symbolic distillation involves extracting symbolic knowledge and representations from LLMs, which can then be used to generate open source data. This open data can serve as the foundation for training new models that are open source, thereby democratizing access to advanced language models. By transitioning from closed source to open source, we can promote transparency, collaboration, and innovation in the field of NLP, aligning with the principles of open science and open AI.

\subsection{Self Improvement of LLMs}
Reinforcement Learning from Human Feedback (RLHF) has emerged as a prevalent method for refining LLMs. However, the involvement of human input inherently constrains its efficacy and outcomes to the limits of human capabilities. Upon undergoing fine-tuning, LLMs can surpass human performance levels. Leveraging these enhanced models to autonomously fine-tune themselves, either via rewards\cite{yuan2024self} or prompt tuning or alternative mechanisms, presents a viable strategy for eliminating the limitations imposed by human intervention opening the gateway for Superintelligence. When employing Reinforcement Learning (RL) for fine-tuning LLMs by themselves, opting for Neurosymbolic RL approaches is often more advantageous. This is because Neurosymbolic RL not only aids in the tuning process but also enhances the model with the ability to interpret and explain its decision-making process comprehensively\cite{acharya2023neurosymbolic}.

\subsection{Cross-domain Symbiosis}
Symbolic Knowledge extracted from LLMs extends its utility beyond the linguistic domain. Studies, such \cite{park2023localized}, demonstrate that textual knowledge can augment visual models by offering explanations and enhancing efficiency. This interdisciplinary application can be further leveraged in diverse fields such as medical imaging, autonomous driving, and surveillance, serving not only to elucidate model outputs but also to improve transfer from one domain to another(simulation to real) by providing the semantic anchors\cite{velasquez2023darpa}. This cross-domain synergy highlights the potential of Symbolic Knowledge in broadening the applicability and understanding of complex AI systems.

\subsection{Industrial Applications}
Symbolic knowledge distillation reveals a critical insight: the effectiveness of LLMs is significantly influenced not only by their size (number of parameters) but more importantly by the quality of the datasets on which they are trained. It highlights the significant role of symbolic knowledge distillation in enhancing domain-specific AI applications by fine-tuning LLMs with specialized corpora and instruction-following data. Notable implementations include LawyerLLaMA\cite{huang2023lawyer} and LawGPT\cite{cui2023chatlaw} for legal services, HuatuoGPT\cite{zhang2023huatuogpt} and ChatDoctor\cite{li2023chatdoctor} for medical applications, XuanYuan\cite{zhang2023xuanyuan} for finance, DARWIN Series\cite{xie2023darwin} and SciGLM\cite{zhang2024sciglm} for scientific research. These tailored models demonstrate substantial improvements in accuracy, efficiency, and usability, showcasing the transformative potential of symbolic knowledge distillation in various industries.

\section{Challenges}
\subsection{Ensuring Data Quality and Diversity in Datasets} While symbolic knowledge distillation from LLMs promises to enhance dataset quality, a major challenge is ensuring the high quality and representativeness of the generated data. The datasets derived from LLMs may inherit biases or inaccuracies present in the original training data of these models. This can lead to the propagation of errors and skewed perspectives in the new datasets, affecting the reliability and fairness of AI systems trained on them. Ensuring data quality requires rigorous validation processes and mechanisms to identify and mitigate biases, which can be resource-intensive, complex, is still an not so explored area.

\subsection{Balancing Automation and Human Oversight in Dataset Generation} While utilizing machines under human guidance can reduce costs, achieving the right balance between automation and human oversight is challenging. Over-reliance on automation may lead to oversight of nuanced or exceptional cases that require human judgment. Conversely, excessive human intervention can negate the efficiency gains from automation. Establishing effective protocols and systems for human-machine collaboration, where machines handle routine tasks while humans oversee and intervene as needed, is crucial but difficult to optimize.

\subsection{Developing Compact Models Without Compromising Performance} Creating smaller models from LLMs that maintain high performance levels is a significant challenge.There are research efforts to quantize LLMs to ultra-low bit sizes, their performance has been found lacking and does not meet the standards required for industrial applications\cite{shao2023omniquant}\cite{shang2023pb}. Symbolic Knowledge Distillation has shown promise in specific, narrower fields such as translation, summarization, and commonsense reasoning. However, it must evolve into a comprehensive symbolic knowledge base capable of generalizing across all domains. Developing these compact models requires sophisticated techniques to compress and optimize the knowledge transfer without losing the nuances and depth of the original model.

\subsection{Effective Instruction Tuning for Diverse Applications} Instruction tuning in AI models poses the challenge of adapting to a wide range of instructions and use cases. Models must be versatile enough to understand and execute a variety of commands accurately across different domains and contexts. This requires extensive training and fine-tuning, which can be resource-intensive. Moreover, ensuring that the models remain adaptable and up-to-date with evolving user needs and language usage is an ongoing challenge, necessitating continuous monitoring and updates.

\subsection{  Adaptability and Continuous Learning}   Ensuring that distilled models can adapt to new information and evolving data landscapes is challenging. Continuous learning mechanisms that allow models to update their knowledge without compromising efficiency or requiring complete retraining are essential for keeping distilled models relevant and effective.

\section{  Lesson Learned and Key Takeaways}
  \subsection{Efficiency Through Distillation} Symbolic knowledge distillation demonstrates a powerful method to enhance the efficiency of LLMs. By distilling complex, large-scale models into smaller, more manageable versions without significant loss in performance, researchers can achieve remarkable efficiency gains. This approach not only reduces computational requirements but also makes advanced AI capabilities more accessible for applications with limited resources.

  \subsection{Advancement in Commonsense Reasoning} The transition to a 'from-machine-to-corpus-to-machine' paradigm marks a significant advancement in commonsense reasoning. This innovative approach, through the creation of extensive and diverse datasets like ATOMIC and models like NOVACOMET, underscores the potential of machine-generated knowledge in improving AI's understanding and application of commonsense knowledge.

  \subsection{Innovation in Data Generation and Use by Collaborating Human Intelligence and Machine Capabilities} LLMs has the potential in generating high-quality, diverse datasets. These datasets serve as a foundation for training more robust models, emphasizing the importance of data quality, diversity, and the innovative use of symbolic knowledge in dataset creation. The effective collaboration between human oversight and automated processes in dataset generation and task execution highlights the synergistic potential of combining human intelligence with machine efficiency. This collaboration is key to overcoming current limitations and unlocking new capabilities in AI systems.

  \subsection{Cross-Domain Applications} The applications of symbolic knowledge distillation extend beyond NLP into areas such as visual commonsense reasoning and mathematical proof solving. This cross-domain applicability showcases the versatility of distilled models and their potential to revolutionize various fields by enhancing model performance and understanding.

  \subsection{Instruction Tuning and Generation} The development and refinement of techniques for instruction tuning and generation signify a leap towards creating more user-friendly and intuitive AI systems. Models capable of generating their own instructions or being finely tuned to understand and execute specific commands can lead to more natural and effective human-AI interactions.

  \subsection{Challenges and Opportunities} While the advancements are notable, they also underscore challenges such as ensuring data quality, balancing automation with human oversight, and developing compact models without compromising performance. Addressing these challenges presents opportunities for further research and innovation in model training, dataset creation, and the development of algorithms for enhanced capabilities and benchmark for the evaluation.
  
 To address the identified gaps in current research on symbolic knowledge distillation, it is crucial to first ensure the quality and diversity of datasets through rigorous validation to identify and mitigate biases inherited from LLMs, ensuring the trustworthy knowledge distillation. Balancing automation and human oversight is also essential; effective protocols for human-machine collaboration can optimize efficiency while ensuring nuanced cases are handled appropriately.Though the size of data required for efficient distillation is still unknown, research\cite{zhou2024lima} propose that only 1000 high quality human curated data is enough. Another challenge is developing compact models without compromising performance, which requires sophisticated techniques to compress and optimize knowledge transfer while maintaining the depth of the original models. Effective instruction tuning for diverse applications demands extensive training and fine-tuning to ensure models can accurately execute various commands across domains. Ensuring adaptability and continuous learning in distilled models is vital, necessitating mechanisms for ongoing updates without compromising efficiency. Addressing these areas will advance symbolic knowledge distillation towards more reliable and practical applications.

\section{Conclusion}
This survey paper has explored the emerging and crucial domain of symbolic knowledge distillation in LLMs. As LLMs continue to grow in scale and complexity, the need to effectively extract and represent their extensive knowledge becomes paramount. By categorizing existing research based on methodologies and applications, we have highlighted how symbolic knowledge distillation can enhance the transparency and functionality of smaller, more efficient AI models. This comprehensive overview underscores the significance of symbolic knowledge distillation in advancing more accessible and efficient AI systems. While there is a notable lack of comprehensive research in this area, our survey paper fills this crucial gap by offering an extensive review of the current state of symbolic knowledge distillation in LLMs, shedding light on methodologies, challenges, and advancements in this field.

\bibliographystyle{IEEEtran}
\bibliography{ref.bib}

\begin{IEEEbiography}[{\includegraphics[width=1in,height=1.25in,,clip,keepaspectratio]{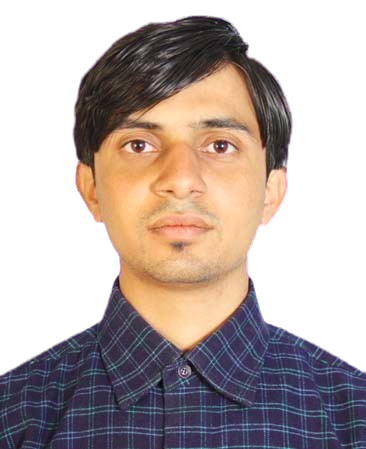}}]{Kamal Acharya}{\space}(Graduate Student Member, IEEE) received his Engineering degree in Electronics and Communication Engineering from Tribhuvan University, Kathmandu, Nepal in 2011 and Masters degree in Information System Engineering from Purbanchal University, Kathmandu, Nepal in 2019. Currently, he is pursuing PhD. in the Information Systems from University of Maryland, Baltimore County (UMBC), Baltimore, MD.

He has been involved in teaching profession for about 7 years in the various universities of Nepal, Tribhuvan University and Purbanchal Univesity were among few of them. He is mainly associated with the courses like programming(C,C++,Python), Computer Networks and Computer Architecture. He is working as Graduate Research Assistant in UMBC. He is also serving as an reviewer for IEEE Transactions on Artificial Intelligence, IEEE Transactions on Intelligent Transportation Systems and IEEE SMC Magazine. His preferred areas of research are Natural Language Processing(NLP), Deep Learning and Reinforcement Learning.
\end{IEEEbiography}

\begin{IEEEbiography}
[{\includegraphics[width=1in,height=1.25in,,clip,keepaspectratio]{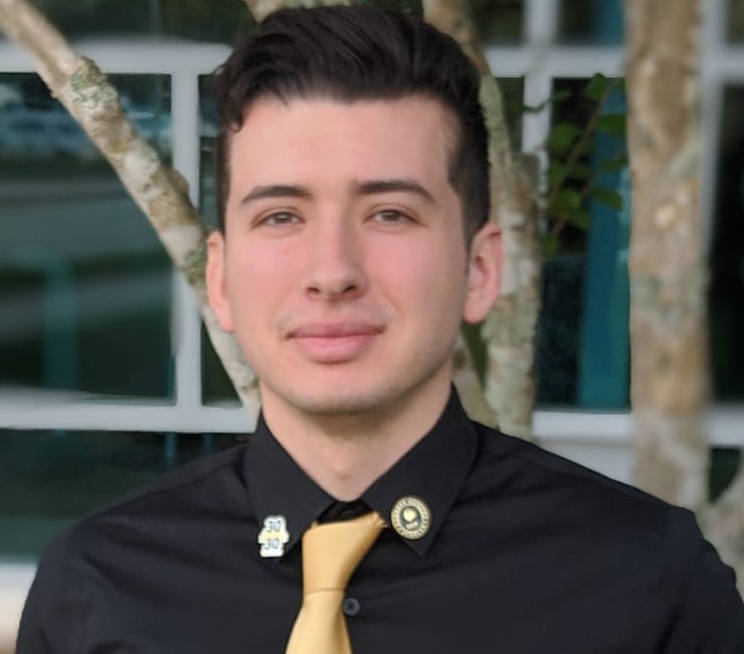}}]{Alvaro Velasquez} is a program manager in the Innovation Information Office (I2O) of the Defense Advanced Research Projects Agency (DARPA), where he currently leads the Assured Neuro-Symbolic Learning and Reasoning (ANSR) program. Before that, Alvaro oversaw the machine intelligence portfolio of investments for the Information Directorate of the Air Force Research Laboratory (AFRL). Alvaro received his PhD in Computer Science from the University of Central Florida in 2018 and is a recipient of the AAAI Distinguished Paper Award, the National Science Foundation Graduate Research Fellowship Program (NSF GRFP) award, the University of Central Florida 30 Under 30 award, and best paper and patent awards from AFRL. He has co-authored over 80 papers and two patents and serves as Associate Editor of the IEEE Transactions on Artificial Intelligence.
\end{IEEEbiography}

\begin{IEEEbiography}
[{\includegraphics[width=1in,height=1.25in,,clip,keepaspectratio]{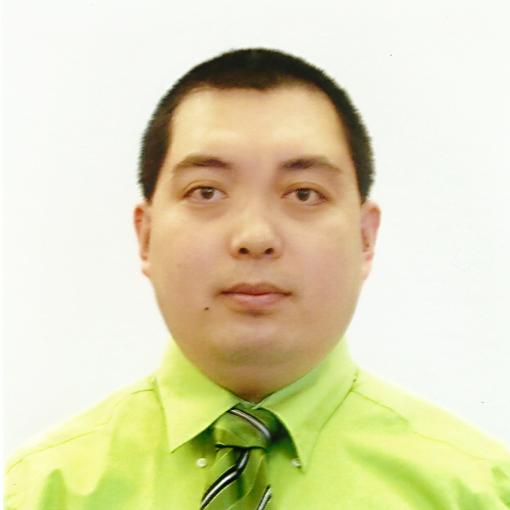}}]{Houbing Herbert Song} (M'12–SM'14-F'23) received the Ph.D. degree in electrical engineering from the University of Virginia, Charlottesville, VA, in August 2012.

He is currently a Professor, the Founding Director of the NSF Center for Aviation Big Data Analytics (Planning), the Associate Director for Leadership of the DOT Transportation Cybersecurity Center for Advanced Research and Education (Tier 1 Center), and the Director of the Security and Optimization for Networked Globe Laboratory (SONG Lab, www.SONGLab.us), University of Maryland, Baltimore County (UMBC), Baltimore, MD. He is a Distinguished Visiting Fellow of the Scottish Informatics and Computer Science Alliance (SICSA). Prior to joining UMBC, he was a Tenured Associate Professor of Electrical Engineering and Computer Science at Embry-Riddle Aeronautical University, Daytona Beach, FL. He serves as an Associate Editor for IEEE Transactions on Artificial Intelligence (TAI) (2023-present), IEEE Internet of Things Journal (2020-present), IEEE Transactions on Intelligent Transportation Systems (2021-present), and IEEE Journal on Miniaturization for Air and Space Systems (J-MASS) (2020-present). He was an Associate Technical Editor for IEEE Communications Magazine (2017-2020). He is the editor of ten books, the author of more than 100 articles and the inventor of 2 patents. His research interests include AI/machine learning/big data analytics, cyber-physical systems/internet of things, and cybersecurity and privacy. His research has been sponsored by federal agencies (including National Science Foundation, National Aeronautics and Space Administration, US Department of Transportation, and Federal Aviation Administration, among others) and industry. His research has been featured on popular news media outlets, including IEEE Spectrum, IEEE GlobalSpec's Engineering360, IEEE Transmitter, insideBIGDATA, Association for Uncrewed Vehicle Systems International (AUVSI), Security Magazine, CXOTech Magazine, Fox News, U.S. News \& World Report, The Washington Times, and New Atlas. 

Dr. Song is an IEEE Fellow, an Asia-Pacific Artificial Intelligence Association (AAIA) Fellow, an ACM Distinguished Member, and a Full Member of Sigma Xi. Dr. Song has been a Highly Cited Researcher identified by Web of Science since 2021. He is an ACM Distinguished Speaker (2020-present), an IEEE Computer Society Distinguished Visitor (2024-present), an IEEE Communications Society (ComSoc) Distinguished Lecturer (2024-present), an IEEE Intelligent Transportation Systems Society (ITSS) Distinguished Lecturer (2024-present), an IEEE Vehicular Technology Society (VTS) Distinguished Lecturer (2023-present) and an IEEE Systems Council Distinguished Lecturer (2023-present). Dr. Song received Research.com Rising Star of Science Award in 2022, 2021 Harry Rowe Mimno Award bestowed by IEEE Aerospace and Electronic Systems Society, and 10+ Best Paper Awards from major international conferences, including IEEE CPSCom-2019, IEEE ICII 2019, IEEE/AIAA ICNS 2019, IEEE CBDCom 2020, WASA 2020, AIAA/ IEEE DASC 2021, IEEE GLOBECOM 2021 and IEEE INFOCOM 2022. He has been an IEEE Impact Creator since 2023.

\end{IEEEbiography}

\end{document}